\documentclass[sigconf]{acmart}

\usepackage{balance}

\def\BibTeX{{\rm B\kern-.05em{\sc i\kern-.025em b}\kern-.08emT\kern-.1667em\lower.7ex\hbox{E}\kern-.125emX}}
    
\copyrightyear{2019}
\acmYear{2019}
\acmConference[MM '19]{Proceedings of the 27th ACM International Conference on
Multimedia}{October 21--25, 2019}{Nice, France}
\acmBooktitle{Proceedings of the 27th ACM International Conference on Multimedia (MM '19),
October 21--25, 2019, Nice, France}
\acmPrice{15.00}
\acmDOI{10.1145/3343031.3351092}
\acmISBN{978-1-4503-6889-6/19/10}

\usepackage{multirow}
\usepackage{cmath}

\begin{document}

\fancyhead{}

\title{Domain-Specific Embedding Network for Zero-Shot Recognition}

\author{Shaobo Min}
\email{mbobo@mail.ustc.edu.cn}
\affiliation{%
  \institution{University of Science and Technology of China}
}

\author{Hantao Yao}
\email{hantao.yao@nlpr.ia.ac.cn}
\affiliation{%
  \institution{National Laboratory of Pattern Recognition, Institute of Automation, Chinese Academy of Sciences}
  }

\author{Hongtao Xie}
\authornote{Corresponding Author.}
\email{htxie@ustc.edu.cn}
\affiliation{%
  \institution{University of Science and Technology of China}
}

\author{Zheng-Jun Zha}
\email{zhazj@ustc.edu.cn}
\affiliation{%
  \institution{University of Science and Technology of China}
}

\author{Yongdong Zhang}
\email{zhyd73@ustc.edu.cn}
\affiliation{%
  \institution{University of Science and Technology of China}
}

%
\renewcommand{\shortauthors}{Trovato and Tobin, et al.}

%
\begin{abstract}
Zero-Shot Learning (ZSL) seeks to recognize a sample from either seen or unseen domain by projecting the image data and semantic labels into a joint embedding space.
However, most existing methods directly adapt a well-trained projection from one domain to another, thereby ignoring the serious bias problem caused by domain differences. 
To address this issue, we propose a novel Domain-Specific Embedding Network (DSEN) that can apply specific projections to different domains for unbiased embedding, as well as several domain constraints. In contrast to previous methods, the DSEN decomposes the domain-shared projection function into one domain-invariant and two domain-specific sub-functions to explore the similarities and differences between two domains.
To prevent the two specific projections from breaking the semantic relationship, a semantic reconstruction constraint is proposed by applying the same decoder function to them in a cycle consistency way.
Furthermore, a domain division constraint is developed to directly penalize the margin between real and pseudo image features in respective seen and unseen domains, which can enlarge the inter-domain difference of visual features.
Extensive experiments on four public benchmarks demonstrate the effectiveness of DSEN with an average of $9.2\%$ improvement in terms of harmonic mean.
The code is available in \url{https://github.com/mboboGO/DSEN-for-GZSL}.
\end{abstract}

%
%


%

%

%
\maketitle
\begin{figure}[t]
	\begin{center}
		\includegraphics[width=0.95\linewidth]{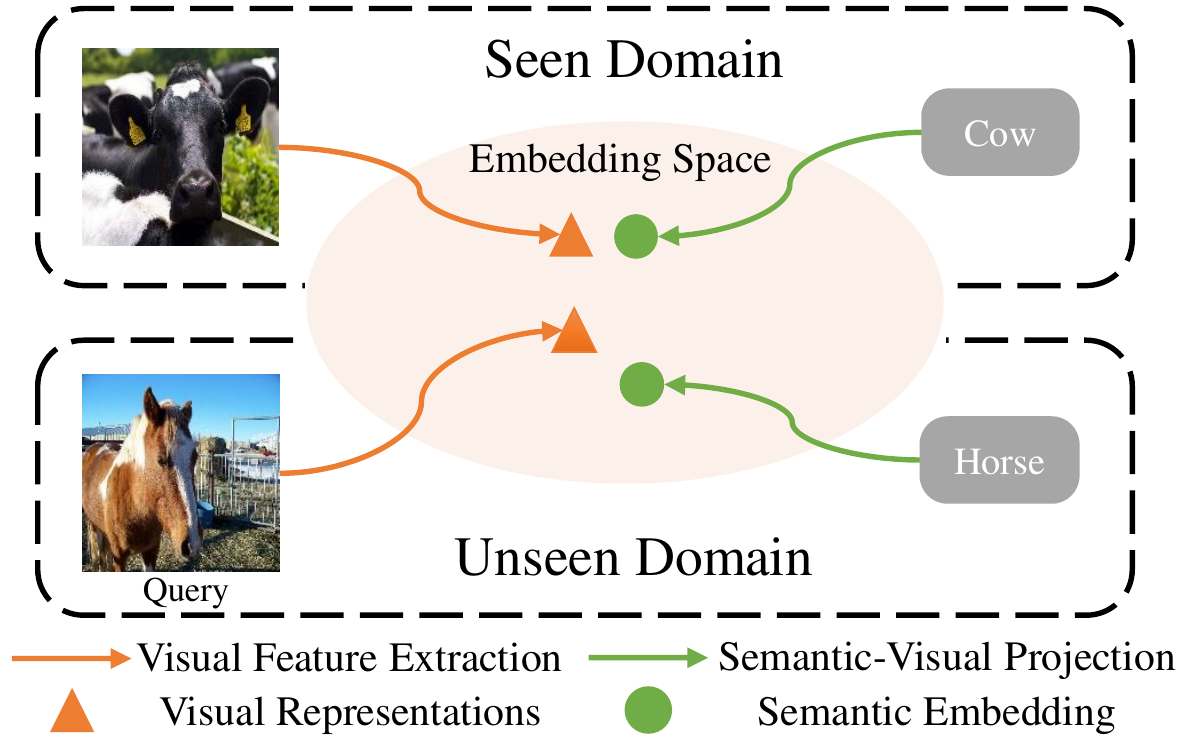}
	\end{center}
	\caption{A diagram of generalized zero-shot recognition, which associates images and semantic labels in a joint embedding space.}
	\label{fig:task}
\end{figure}

\section{Introduction}
\label{sec:intro}
Traditional recognition tasks has progressed with the help of massive labeled images and deep models \cite{Simonyan2014,He2016,he2018only,fang2018attention,zheng2018fast,wang2018multimodal,xie2019automated,CPGBN_ICML2019}.
However, their major disadvantage is that they cannot recognize the images belonging to unseen categories, and it is laborious to collect sufficient labeled images for various tasks. To tackle this problem, generalized Zero-Shot Learning (ZSL) \cite{Palatucci2009,Akata2013,Lampert2014,Romera-Paredes2015a,yang2016zero,Morgado2017,Xian2018,long2018pseudo} has attracted a lot of attention in recent years.
A generalized zero-shot recognition is defined as recognizing a sample from either seen or unseen domain, which contains disjoint categories.
A general paradigm is to project the image data and semantic labels,~\emph{e.g.,} category attributes~\cite{Farhadi2009,Lampert2009}, into a joint embedding space, where recognition becomes a nearest neighbor searching problem~\cite{Song2018}. 
A visual diagram is shown in Figure~\ref{fig:task}.
The major challenge of this paradigm is that the different data distributions between two domains lead to serious domain shift problems \cite{Fu2014,Kodirov2015}, which make the embedding features biased towards the seen domain.

\begin{figure*}[t]
	\begin{center}
		\includegraphics[width=0.9\linewidth]{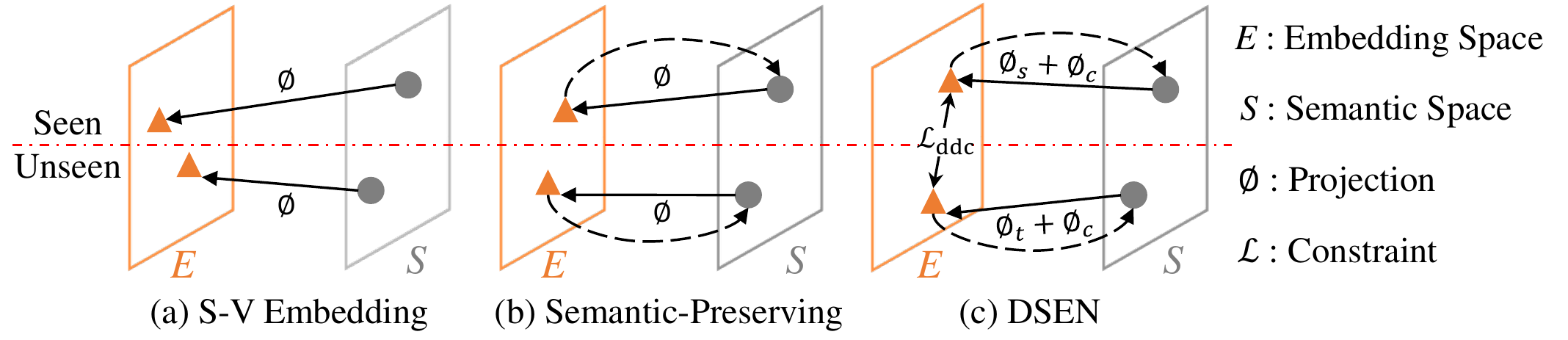}
	\end{center}
	\caption{Comparison of DSEN with related GZSL paradigms. a) The embedding space is spanned by visual features. b) Preserving the category relationship in the embedding space. c) the proposed DSEN, which introduces two extra $\phi_s$ and $\phi_t$ to model domain-specific knowledge and a domain division constraint $\mathcal{L}_{ddc}$ to enlarge the inter-domain difference.}
	\label{fig:rw}
\end{figure*}

To address this issue, existing methods focus on learning a robust projection between visual representations and semantic labels.
The related methods can be coarsely classified into two classes: embedding-based framework and semantic-preserving framework. 
The embedding-based framework \cite{Tomasev2014,Romera-Paredes2015a,Akata2016,Jiang2017,Verma2018} aims to establish a discriminative embedding space that is shared across two domains.
Two commonly used embedding space is spanned by visual representations \cite{zhang2017learning} and semantic labels \cite{Verma2018}, respectively.
Taking Figure~\ref{fig:rw} (a) as an example, the semantic labels are projected into the visual space to match with the corresponding visual representations, which has been proved robust to Hubness problems \cite{Tomasev2014,Lazaridou2015}.
Different from the embedding-based framework, the semantic-preserving framework \cite{kodirov2017semantic,Annadani2018,Chen2018,Xian2018} focuses on preserving the semantic prototype in an embedding space through an auto-encoder architecture.
An example is shown in Figure~\ref{fig:rw} (b).
Although the above methods are effective for zero-shot problems, they all employ a single shared-projection for both seen and unseen domains.
Due to the broad domain gap, a shared projection function cannot model the full specialty of each domain, leading to biased recognition problem.

In this paper, we propose a novel Domain-Specific Embedding Network (DSEN) to alleviate the domain shift problem in ZSL by applying specific projections to different domains, as well as several domain constraints.
The novelties of DSEN over the previous methods are shown in Figure~\ref{fig:rw} (c).
Instead of a single shared projection, DSEN decomposes the projection function into three components: domain-invariant projection $\phi_c$, seen domain-specific projection $\phi_s$, and unseen domain-specific projection $\phi_t$.
The $\phi_c$ targets to capture the common projection knowledge between two domains, and $\phi_s$ and $\phi_t$ are used to capture the domain-specific projection knowledge. 
Notably, $\phi_s$ and $\phi_t$ should project the semantic labels of different domains into a shared embedding space $E$ in Figure~\ref{fig:rw} (c) for cross-domain recognition.
To this end, a semantic reconstruction constraint is designed, by applying the same decoder function to both $\phi_s$ and $\phi_t$ in a cycle consistency way, to preserve the shared semantic relationship in $E$.
Compared to using a single shared projection, our domain-specific projections can generate less biased embedding features due to domain specialty modeling.

Furthermore, a domain division constraint is developed to enlarge both intra- and inter-domain discrimination of visual features, based on pseudo visual data in the unseen domain.
Besides fully supervised learning in the seen domain, our domain division constraint restricts the noisy pseudo data to have a uniform label distribution in seen categories.
The advantages of this constraint mechanism are as follows: a) being robust to the noisy pseudo visual features; b) directly enlarging the visual margin between two domains; and c) can be trained end-to-end.
Consequently, the decision boundary of visual features between two domains becomes more clear, which allows us to utilize specific classifiers in determinate searching space for unbiased recognitions.

Our contributions are threefold:
\begin{itemize}

\item We propose a novel Domain-Specific Embedding Network (DSEN) by applying specific projections to two domains, which can better capture domain similarities and differences for unbiased embedding.

\item A domain division constraint is designed to effectively enhance both intra- and inter-domain discrimination, based on real and pseudo visual data in two domains. Besides, it also enables DSEN to be trained end-to-end.

\item The proposed DSEN obtains the sate-of-the-art performance on four public datasets with an average of 9.2\% improvement in terms of harmonic mean.

\end{itemize}

\section{Related Work}
Three types of related techniques are discussed in this section.
\subsection{Embedding-based Zero-Shot Learning}
A general paradigm of zero-shot recognition targets to project the image representations and semantic labels into a joint embedding space, where the recognition becomes a nearest neighbor searching problem \cite{Song2018}.
This process is called as an embedding-based method, which is one of the most popular ZSL strategies.
As the seen and unseen domains have disjointed categories, the additional semantic information, such as attributes \cite{Farhadi2009,Lampert2009} and word vectors \cite{Pennington2014,Niu2017}, are used to construct a relationship between these two domains.
Among these methods, Frome~\emph{et al.}~\cite{Frome2013} and Akata~\emph{et~al.}~\cite{Akata2016} use the bilinear embedding model trained with a pairwise ranking loss.
The ESZSL model \cite{Romera-Paredes2015a} constructs an embedding space with a Frobenius norm regularization, and Qiao~\emph{et~al.}~\cite{Qiao2016} extend this work to online documents by suppressing the noise with an extra $l_{1,2}$ norm.
In addition, Akata~\emph{et al.}~\cite{Akata2015} build a joint embedding space with several compatibility functions, which is improved in \cite{Xian2016} by incorporating latent variables. 
Zhang~\emph{et al.}~\cite{Zhang2018} employ a non-linear kernel to generate a mapping between visual representations and attributes. 
In spite of the promising performance, the above methods directly project the visual representations into space spanned by semantic labels, which suffer from Hubness problems \cite{Radovanovic2010,Tomasev2014,Lazaridou2015}.
The Hubness problem is defined as a few points being the nearest neighbors of most of the other points, which is caused by that projecting a visual feature with high dimensions into an attributes space with low dimensions shrinks the variance of the projected data points \cite{zhang2017learning}.
Therefore, a few methods \cite{Shigeto2015,zhang2017learning,Xian2018} use an embedding space spanned by visual features, which is defined as a semantic-visual embedding.
Although the previous methods are effective, insufficient semantic embedding limits their further applications due to serious domain shift problems.
For example, a testing sample from an unseen domain tends to be recognized from one of the seen categories.

\subsection{Semantic-Preserving Framework}
To alleviate the domain shift problems, many recent works target to preserve the semantic prototype in an embedding space.
The motivation under this tactic is that the semantic prototype is robust to domain change, which is beneficial to train a robust projection function.
Among these methods, SAE \cite{kodirov2017semantic} and SP\_AEN~\cite{Chen2018} use an auto-encoder architecture on the embedded space to make their embedding features discriminative.
Jiang~\emph{et al.}~\cite{Jiang_2018_ECCV} propose a coupled dictionary learning model to preserve the visual-semantic structures with semantic prototypes.
Especially, Annadani~\emph{et al.}~\cite{Annadani2018} preserve the semantic relationships in visual space by decomposing the relation between categories into three groups.
Although the embedding-based methods and semantic-preserving framework can effectively solve the ZSL problems, they are mainly based on using a single shared projection across two domains, which ignore the large domain differences.

\subsection{Synthetic Data-Based Methods}
Recently, synthetic data-based methods \cite{bucher2017generating,Mishra_2018_CVPR_Workshops,long2018pseudo,Verma2018,xian2018feature} have been proposed, and they have obtained state-of-the-art performance.
In contrast to embedding-based methods, they train a softmax classifier with full supervision on the union of real and synthetic visual data from seen and unseen domains.
The synthetic visual data is obtained by a specific generator, such as GAN \cite{Goodfellow2014} and its variants, based on the unseen domain attributes.
Consequently, their models are more robust to domain shifts than embedding-based methods, based on both seen and unseen visual features.
However, fully-supervised learning is sensitive to noisy synthetic visual data, which has not been fully exploited.

\begin{figure*}[t]
	\begin{center}
		\includegraphics[width=0.88\linewidth]{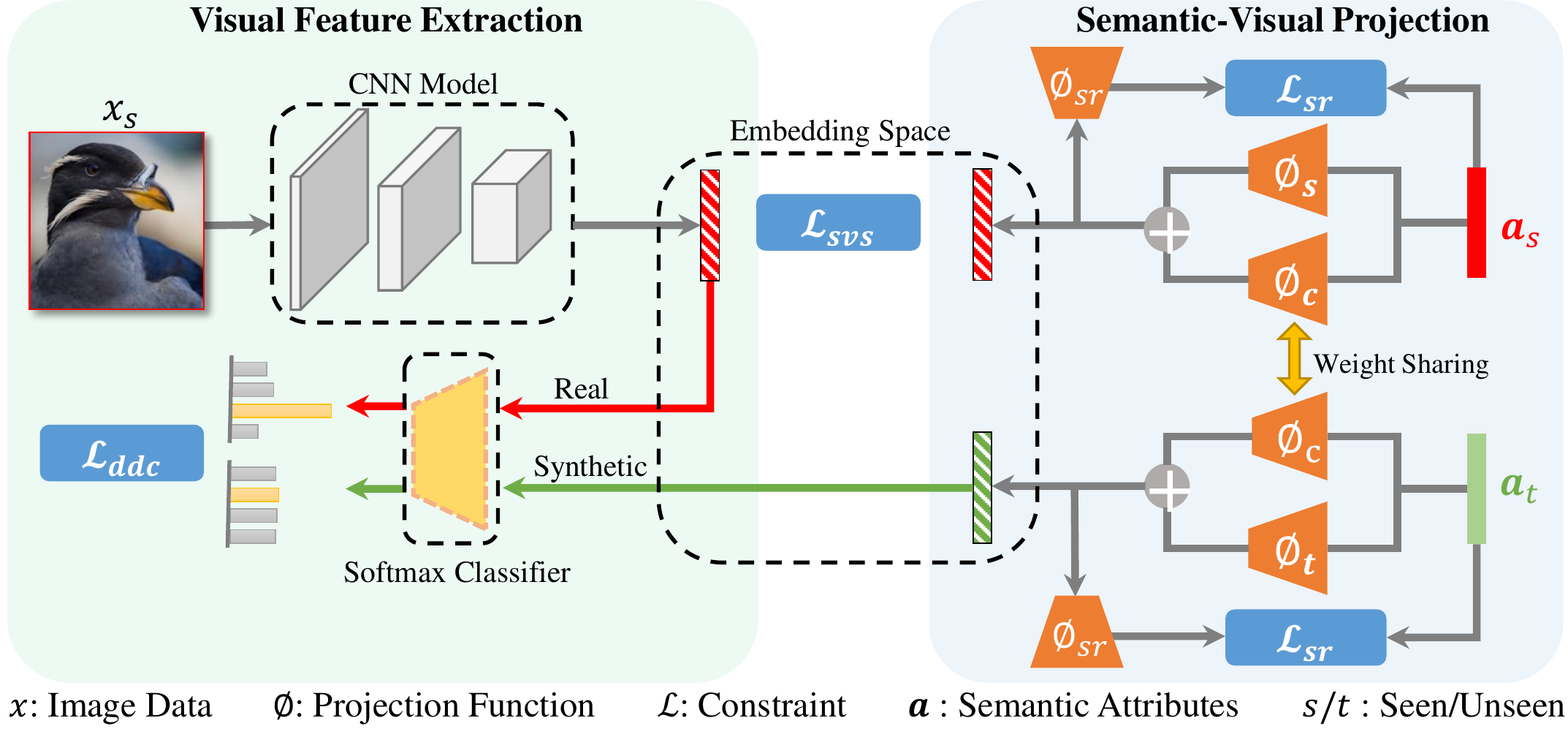}
	\end{center}
	\caption{The pipeline of training Domain-Specific Embedding Network. Besides the domain-shared projection $\phi_c$, DSEN trains two extra domain-specific $\phi_s$ and $\phi_t$ to better capture domain specialties. Furthermore, the domain division constraint $\mathcal{L}_{ddc}$ makes the visual embedding features in two domains distinguishable. The whole network is trained end-to-end.}
	\label{fig:pipeline}
\end{figure*}
\section{Domain-Specific Embedding Network}
\label{sec:method}
We first describe the problem formulation of the Domain-Specific Embedding Network in Sec.~\ref{sec:pf} and then provide a detailed implementation of DSEN. The pipeline is shown in Figure~\ref{fig:pipeline}. 

\subsection{Problem Formulation}
\label{sec:pf}
Let $\mathcal{S}=\{(x_{s},y_{s},\boldsymbol{a}_{s})|x_{s}\in\mathcal{X}_s, y_{s}\in\mathcal{Y}_s,\boldsymbol{a}_{s}\in\mathcal{A}_s\}$ represents the seen domain dataset, where $y_{s}$ and $\boldsymbol{a}_{s}$ are the class labels and semantic attributes for each image $x_{s}$, respectively. $\mathcal{T}=\{(x_{t},y_{t},\boldsymbol{a}_{t})|x_{t}\in\mathcal{X}_t, y_{t}\in\mathcal{Y}_t,\boldsymbol{a}_{t}\in\mathcal{A}_t\}$ is similarly defined as the unseen domain dataset, where $\mathcal{Y}_s\cap \mathcal{Y}_t = \varnothing$.
Given the seen domain data $\mathcal{S}$ and unseen domain labels $\mathcal{Y}_t$ with attributes $\mathcal{A}_t$, the target of a generalized ZSL task is to recognize an image from either $\mathcal{X}_s$ or $\mathcal{X}_t$. 

Based on the above definition, a basic objective for our DSEN is to:
\begin{eqnarray}
\min_{W_{\phi}}\sum_{x_s\in\mathcal{X}_s}d(f(x_s),\phi(\boldsymbol{a}_s)),
\label{eq:obj}
\end{eqnarray}
where $f(\cdot)$ is the visual feature extractor for visual images.
$\phi(\cdot)$ is a semantic-visual projection function with trainable weights $W_{\phi}$.
Notably, $\phi(\boldsymbol{a}_s)$ is the semantic embedding.
The distance function $d(\cdot)$ computes the negative cosine distance between two features $\boldsymbol{v}_1$ and $\boldsymbol{v}_2$ by:
\begin{eqnarray}
d(\boldsymbol{v_{1}},\boldsymbol{v_{2}})=-\frac{<\boldsymbol{v_{1}},\boldsymbol{v_{2}}>}{||\boldsymbol{v_{1}}||_2||\boldsymbol{v_{2}}||_2}.
\label{eq:s}
\end{eqnarray}

In most existing methods~\cite{Shigeto2015,zhang2017learning,Annadani2018}, $\phi$ is trained on $\mathcal{S}$ and directly adapted to $\mathcal{T}$ and $f(\cdot)$ is fixed by using pre-trained visual feature extractor.

\subsection{Domain-Specific Projections}
One leading cause of the domain shift problem is that a shared $\phi_c$ cannot model the full differences between two domains, thereby making generated embedding features towards the seen domain. 
Targeting to model the differences between two domains, we decompose the projection function into three parts, which are domain-invariant projection $\phi_c$, seen domain-specific projection $\phi_{s}$, and unseen domain-specific projection $\phi_{t}$.
Thus, the embedding features from seen and unseen domains become the combination of two sub-features:
\begin{eqnarray}\label{eq:DSP}
\phi(\boldsymbol{a})=\left\{
\begin{aligned}
\phi_s(\boldsymbol{a})+\phi_c(\boldsymbol{a}) \qquad if~ \boldsymbol{a}\in\mathcal{A}_s, \\
\phi_t(\boldsymbol{a})+\phi_c(\boldsymbol{a}) \qquad if~ \boldsymbol{a}\in\mathcal{A}_t,
\end{aligned}
\right.
\end{eqnarray}
where $\phi_c$ is used to capture the common knowledge between two domains, and $\phi_s$ and $\phi_t$ capture the specific characteristics for seen and unseen domains, respectively.
Compared to existing methods that use a single shared projection, the additional domain-specific projections can better accommodate domain differences, yielding more discriminative embedding features.
By taking the $\phi_c$ and $\phi_s$ into consideration, Eq.~\eqref{eq:obj} becomes minimizing:
\begin{eqnarray}
\mathcal{L}_{svs}=\sum_{x_s\in\mathcal{X}_s}d(f(x_s),\phi_s(\boldsymbol{a}_s)+\phi_c(\boldsymbol{a}_s)).
\label{eq:obj_svs}
\end{eqnarray}

Different from $\phi_{s}$, $\phi_{t}$ is difficult to train due to unavailable $f(x_t)$. 
Specifically, it is hard to constrain $\boldsymbol{a}_s$ and $\boldsymbol{a}_t$ to be projected into a shared embedding space using different $\phi_s$ and $\phi_t$, for cross domain recognition.
To achieve this goal, a semantic reconstruction constraint $\mathcal{L}_{sr}$ is designed by applying the same decoder function to both $\phi_s+\phi_c$ and $\phi_t+\phi_c$ for semantic label reconstruction.
The motivation is that the semantic labels are shared across two domains; thus $\mathcal{L}_{sr}$ can constrain the semantic embedding from two specific projections to be associated in a shared embedding space.
First, $\phi_t$ is initialized on the well-trained $\phi_s$ for projection knowledge transfer.
Then, the semantic information in $\boldsymbol{a}_s$ and $\boldsymbol{a}_t$ are simultaneously encoded into $\phi_s$ and $\phi_t$ in a semantic cycle consistency way.
Notably, using $\phi_s$ as the initialized $\phi_t$ can facilitate the convergence.
Consequently, $\mathcal{L}_{sr}$ enables $\phi_t$ to capture the effective projection knowledge in the unseen domain based on $\mathcal{A}_t$, which will be illustrated subsequently.

Inspired by the applications of auto-encoding architecture in unsupervised representation learning, a domain-specific auto-encoder architecture is used to encode the semantic information in both $\phi_t$ and $\phi_s$ by:
\begin{eqnarray}\label{eq:L_rec}
\begin{split}
\mathcal{L}_{sr} = &\sum_{\boldsymbol{a}_s\in\mathcal{A}_s}||\phi_{sr}(\phi_s(\boldsymbol{a}_s)+\phi_c(\boldsymbol{a}_s))-\boldsymbol{a}_s||^2_2\\
&+\sum_{\boldsymbol{a}_t\in\mathcal{A}_t}||\phi_{sr}(\phi_t(\boldsymbol{a}_t)+\phi_c(\boldsymbol{a}_t))-\boldsymbol{a}_t||^2_2,
\end{split}
\end{eqnarray}
where $\phi_{sr}$ is a shared decoder function for both domains.
From Eq.~\eqref{eq:L_rec}, $\phi_c$ has access to the semantic information in two domains, which can capture the domain similarity information. 
$\phi_s$ and $\phi_t$ only have access to domain-specific information, thereby rendering them to capture specific characteristics of two domains.

Finally, the objective function for domain-specific projections becomes:
\begin{eqnarray}
\min_{W_{\phi_c},W_{\phi_s},W_{\phi_t},W_{\phi_{sr}}}\mathcal{L}_{svs}+\lambda_1\mathcal{L}_{sr},
\label{eq:obj_dsp}
\end{eqnarray}
where $\lambda_1$ is a hyper-parameter used to balance different constraints.
All the encoders $\phi_c$, $\phi_s$, $\phi_t$ and decoder $\phi_{sr}$ are implemented with two fully connection layers followed by ReLU activation.

Consequently, the domain-specific projections $\phi_s$ and $\phi_t$ assist $\phi_c$ to generate less biased embedding features during semantic-visual projection.
The detailed architecture of our domain-specific projections is shown in Figure.~\ref{fig:pipeline}.

\subsection{Domain Division Constraint}
Based on the embedding features from semantic attributes, we further propose a domain division constraint to make the embedding features between two domains distinguishable.

To achieve this goal, we first generate pseudo visual features from category attributes $\boldsymbol{a}_t$ for the unseen domain.
Especially, we regard $\phi(\boldsymbol{a}_t)$ as pseudo visual features, because $\phi(\boldsymbol{a}_t)$ and $f(x_t)$ have similar distributions based on well-trained semantic-visual projections.
With the real visual features $f(x_s)$ and pseudo visual features $\phi(\boldsymbol{a}_t)$, it is intuitive to train a $|\mathcal{Y}_s\cup\mathcal{Y}_t|$-way softmax classifier, which can recognize visual samples from either seen or unseen domain.
However, $\phi(\boldsymbol{a}_t)$ is usually too noisy for a model to use fully supervised learning, which may deteriorate the model performance in the seen domain.
Therefore, DDC just constraints the noisy pseudo features $\phi(\boldsymbol{a}_t)$ to be far away from the seen categorizes, because $\mathcal{Y}_s$ and $\mathcal{Y}_t$ are disjoint.
Thus, a $|\mathcal{Y}_s|$-way softmax classifier $p$ is trained by minimizing:
\begin{eqnarray}\label{eq:L_cls}
\mathcal{L}_{ddc} = -\sum_{x_s\in\mathcal{X}_s}ln~p_{y*}(f(x_s))+\alpha\sum_{\boldsymbol{a}_t\in\mathcal{A}_t}ln~\hat{p}({\phi}(\boldsymbol{a}_t)),
\end{eqnarray}
where $p_{y}(\cdot)$ is the classification score in terms of the ground truth label $y*$, and $\hat{p}(\cdot)$ is the maximum classification score in $\mathcal{Y}_s$.
The first term in Eq.~\eqref{eq:L_cls} is a general cross-entropy softmax loss in the seen domain.
The second term forces $\phi(\boldsymbol{a}_t)$ to have a uniform label distribution in $\mathcal{Y}_s$, which means that the $\phi(\boldsymbol{a}_t)$ should not be recognized as a seen category.
\begin{table}[h]
\begin{center}
\caption{The details of the experimental datasets. $|\mathcal{Y}_{s}|$ and $|\mathcal{Y}_{t}|$ indicate the class numbers in the seen and unseen domains. The train/val/test indicates the image number of the respective split.} \label{tab:datasets}
\begin{tabular}{lcccccc}
  \hline
  Datasets &Attributes&$|\mathcal{Y}_s|$&$|\mathcal{Y}_t|$&train&val&test \\
  \hline
  \hline
  CUB&312&150&50&7,057&1,764&2,967\\
  SUN&102&645&72&10,320&2,580&1,440\\
  AWA2&85&40&10&23,527&5,882&7,913\\
  aPY&64&20&12&5,932&1,483&7,924\\
  \hline
\end{tabular}
\end{center}
\end{table}
$\alpha$ is a hyper-parameter that is used to balance the training effects between real and pseudo visual features on classifier $p$.

Based on Eq~\eqref{eq:L_cls}, the decision boundary between $f(x_s)$ and $f(x_t)$ can be determined by judging whether the label distribution of an input sample is smooth in $\mathcal{Y}_s$. 
Especially for an input image from the seen domain, $\hat{p}(f(x))$ should be extremely large in terms of the true label. Conversely, the $\hat{p}(f(x))$ should be small, indicating a uniform label distribution for an image from the unseen domain.
Furthermore, since the well-trained classifier $p$ can only do categorization in the seen domain, we employ a ranking-based classifier, which is proposed based on nearest neighbor searching, to those samples that are suspected from the unseen domain.
Thus, the final inference of our DSEN can be expressed by:
\begin{eqnarray}\label{eq:zsl_infer}
\hat{y}=\left\{
\begin{aligned}
&\arg\max_{y\in\mathcal{Y}_s} p_y(f(x))\qquad if~ \hat{p}(f(x))>\tau \\
&\arg\min_{y\in\mathcal{Y}_t} d(f(x),\phi(\boldsymbol{a}_t))\quad else,
\end{aligned}
\right.
\end{eqnarray}
where $x\in\mathcal{X}_s\cup\mathcal{X}_t$, and $\hat{y}$ is the final prediction.
$\tau$ is a threshold to determine the domain of an input sample.

With Eq.~\eqref{eq:zsl_infer}, we can divide the searching space for any samples into two sub-spaces.
Once the samples coming from the unseen domain, the ranking-based classifier is used for recognition, of which the search space has been reduced to the unseen domain.
For samples from the seen domain, the softmax classifier $p$ can directly give the confident category predictions.
By reducing the search space, the recognitions in both domains will be measurably improved, which is attributed to our domain division constraint $\mathcal{L}_{ddc}$.

\subsection{Overall Objective}
Finally, the overall objective function of DSEN becomes:
\begin{eqnarray}
\min_{W_{\phi_c},W_{\phi_s},W_{\phi_t},W_{\phi_{sr}},W_{f}}\mathcal{L}_{svs}+\lambda_1\mathcal{L}_{sr}+ \lambda_2\mathcal{L}_{ddc},
\label{eq:obj_daen}
\end{eqnarray}
where $W_{f}$ is the trainable parameters of visual feature extraction function $f(\cdot)$.
$\lambda_1$ and $\lambda_2$ balance different constraints.
Notably, in many existing ZSL methods \cite{Jiang_2018_ECCV,Chen2018,xian2018feature}, $f(\cdot)$ is fixed across different datasets, leading to a weak visual representation $f(x)$.
Instead, $\mathcal{L}_{ddc}$ enables DSEN to be trained end-to-end with a trainable $f(\cdot)$.
Therefore, both visual representations and embedding features from our DSEN are powerful and discriminative.

\section{Experiments}
In this section, experimental analysis on four benchmarks is given to evaluate the proposed DSEN.

\subsection{Experimental Settings}
\noindent\textbf{Datasets.}
We evaluate the proposed method on four widely used benchmarks: Caltech-USCD Birds-200-2011 (CUB) \cite{Welinder2010}, SUN \cite{Patterson2012}, Animals with Attributes 2 (AwA2) \cite{Xian2018}, and Attribute Pascal and Yahoo (aPY) \cite{Farhadi2009}. All the datasets provide annotated attributes. The newly proposed splits of seen/unseen classes in \cite{Xian2018} are used for fair comparisons, which ensure that the test categories are strictly unseen in the pretrained visual projection network \cite{Russakovsky2015}.
The details of the datasets are listed in Table~\ref{tab:datasets}.

\begin{table}
\begin{center}
\caption{The detailed implementations of three baselines.} \label{tab:baselines}
\begin{tabular}{lccc}
  \hline
  Setting&$L_{svs}$&$L_{sr}$&$L_{ddc}$ \\
  \hline
  \hline
  S2V&$\surd$&&\\
  DSP&$\surd$&$\surd$&\\
  DDC&$\surd$&&$\surd$\\
  \hline
  \hline
  DSEN (DSP+DDC)&$\surd$&$\surd$&$\surd$\\
  \hline
\end{tabular}
\end{center}
\end{table}

\noindent\textbf{Implementation details.}
The input images are resized to $480$ along the short side, with data augmentation of $448\times 448$ random cropping and horizontal flipping. 
The visual feature extraction network $f(\cdot)$ is based on the ResNet-101 architecture, which is pre-trained on the ImageNet dataset.
The rest of the networks uses MSRA random initializer \cite{He2016}.
In this work, we employ a two-stage training strategy to train the proposed DSEN. 
It first fixes $f(\cdot)$ and trains the rest with a large learning rate $lr=1\times e^{-3}$, and then it uses a small $lr=1\times e^{-5}$ to train the whole DSEN. The Adam optimizer is used with $\beta=(0.5,0.999)$ and weight decay $5\times e^{-5}$.
For the hyper-parameters in DSEN, we set $\lambda_1=5$ and $\lambda_2=1$ to balance $\mathcal{L}_{svs}$, $\mathcal{L}_{sr}$, and $\mathcal{L}_{ddc}$, and $\alpha=0.1$ in $\mathcal{L}_{ddc}$.
The above hyper-parameter settings are determined according to experiments, and they are applicable to all of our experimental datasets.
$\tau$ will be analyzed in the ablation study.

\noindent\textbf{Evaluation metrics.}
Similar to \cite{Xian2018}, the harmonic mean ($H$) is denoted in Eq.~\eqref{eq:H} to evaluate a model by:
\begin{eqnarray}\label{eq:H}
\begin{split}
H = \frac{2\times \textit{MCA}_t\times \textit{MCA}_s}{\textit{MCA}_t+\textit{MCA}_s},
\end{split}
\end{eqnarray}
where \textit{MCA}$_s$ and \textit{MCA}$_t$ are the Mean Class top-1 Accuracy for the validation (seen) and testing (unseen) sets, respectively.

In the following parts, the experiments are mainly conducted under generalized ZSL settings, where the testing images come from either the seen or unseen domain.

\noindent\textbf{Baselines.}
To demonstrate the effectiveness of different components in DSEN, three baselines are defined:
\begin{itemize}
\item S2V is a general semantic-visual structure with shared projection function $\phi_c$. The visual feature extraction function $f(\cdot)$ is fixed.
\item DSP adds two extra domain-specific projections $\phi_s$ and $\phi_t$ to S2V.
\item DDC applies the domain division constraint $\mathcal{L}_{ddc}$ to S2V, which makes the visual feature extractor $f(\cdot)$ trainable.
\end{itemize}
Finally, DSEN uses both domain-specific projections and $\mathcal{L}_{ddc}$ with a trainable $f(\cdot)$. The details of each baseline are listed in Table~\ref{tab:baselines}.

\begin{figure*}
	\begin{center}
		\includegraphics[width=1\linewidth]{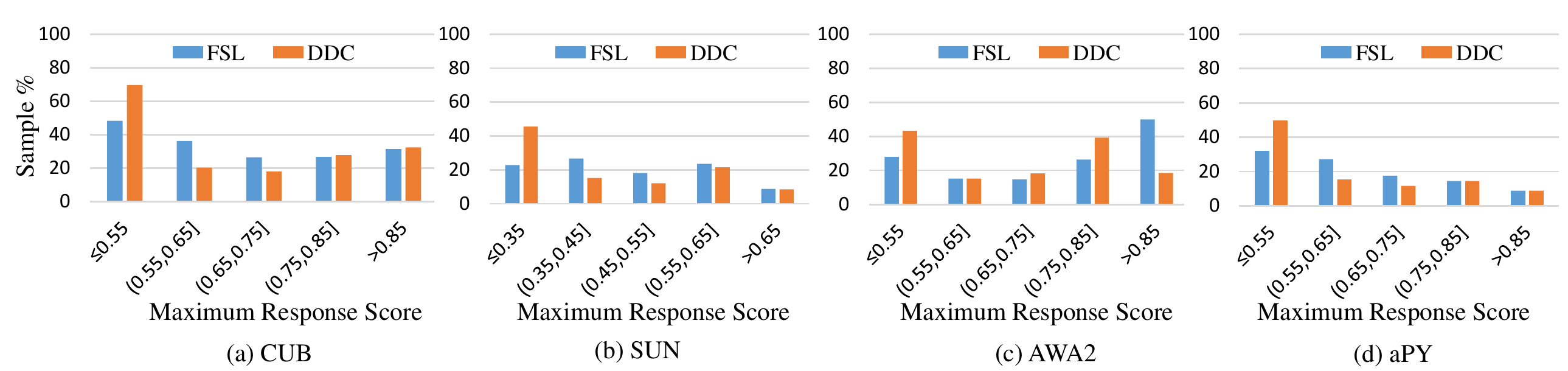}
	\end{center}
	\caption{Distributions of maximum classification score on four datasets. The vertical axis indicates the percentage of unseen domain samples. FSL and DDC represent the fully-supervised learning and our domain division constraint, respectively.}
	\label{fig:c}
\end{figure*}

\begin{table}
\begin{center}
\caption{The effects of each domain-specific projection on CUB.} \label{tab:phi_s_t}
\begin{tabular}{c|ccc|cc}
  \hline
  Baseline&$\phi_c$&$\phi_s$&$\phi_t$&\textit{MCA}$_t$&\textit{MCA}$_s$ \\
    \hline
    \hline
  \multirow{5}{*}{S2V}&$\surd$&&&25.6&56.6\\
  &$\surd$&$\surd$&&27.5&61.9\\
   &$\surd$&&$\surd$&28.3&57.4\\
  &&$\surd$&$\surd$&29.7&60.1\\
  &$\surd$&$\surd$&$\surd$&30.8&62.7\\
  \hline
\end{tabular}
\end{center}
\end{table}


\subsection{Ablation Studies}

\noindent\textbf{Effects of $\phi_c$, $\phi_s$, and $\phi_t$.}
As the domain-specific projections consist of one domain-shared $\phi_c$ and two domain-specific $\phi_s$ and $\phi_t$, we explore their effects by individually applying them to the baseline S2V.
Table~\ref{tab:phi_s_t} shows the results.
From Table~\ref{tab:phi_s_t}, it is observed that applying $\phi_s$ and $\phi_t$ individually to $\phi_c$ yields improvements by $5.3\%$ and $2.7\%$ in terms of $MAC_s$ and $MAC_t$.
This proves that the domain-specific projections effectively capture their characteristic domain information via semantic reconstruction constraint $\mathcal{L}_{sr}$.
Then, we further explore the effects by using totally separated $\phi_s$ and $\phi_t$ without $\phi_c$.
The results show a slight improvement of $1.5\%$ on $MAC_t$.
The reason is that the connection between $\phi_s$ and $\phi_t$ is too weak to guarantee the projected embedding features to be in the same embedding space.
Finally, combining $\phi_s$, $\phi_t$, and $\phi_c$ offers the best performance, which indicates that $\phi_c$ successfully captures the domain similarities during semantic-visual projection.
These experiments prove the effectiveness of domain-specific projections in generating discriminative embedding features.
In addition, random initialization of $\phi_t$ yields a relatively slow convergence speed.


\noindent\textbf{Effects of domain-specific projections.}
As the domain-specific projections play a critical role in the proposed DSEN, we analyze the effect of applying domain-specific projections to different baselines. The related results are summarized in Table~\ref{tab:gzsl}. From Table~\ref{tab:gzsl}, we observe that applying domain-specific projections achieves better performance than using a single shared projection for all datasets, \emph{e.g.,} the DSP and DSEN both achieve $6.0\%$ and $1.9\%$ improvements on the S2V and DDC baselines in terms of $H$ on CUB, respectively. Table~\ref{tab:gzsl} further shows that the domain-specific projections improve the recognition performance on both seen and unseen domains by about $2\%\sim 6\%$ on CUB. 
These achievements demonstrate the effectiveness of the proposed domain-specific projections.

\begin{figure*}
	\begin{center}
		\includegraphics[width=1\linewidth]{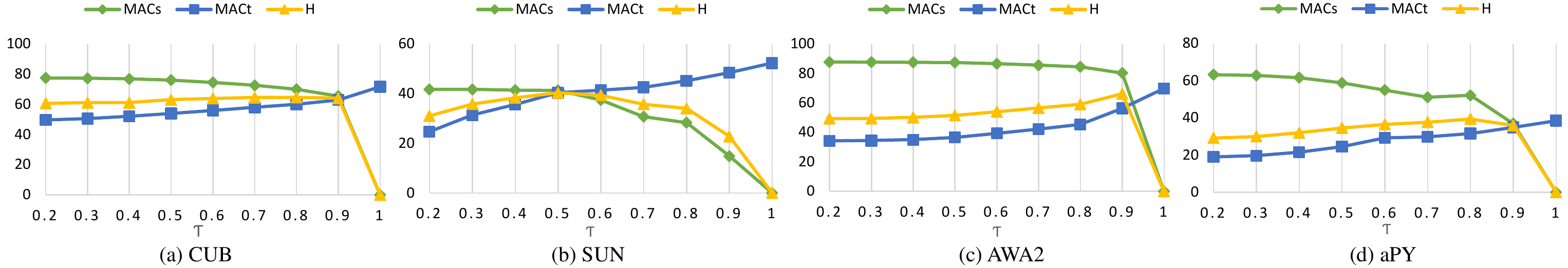}
	\end{center}
	\caption{The performance of DSEN with varying $\tau$ on different datasets. }
	\label{fig:t}
\end{figure*}

\noindent\textbf{Comparison of fully-supervised learning and $\mathcal{L}_{ddc}$.}
We further analyze the superiority of our domain division constraint $\mathcal{L}_{ddc}$ to fully-supervised learning.
The analysis is performed by using noisy pseudo visual data for supervised training.
Given a ZSL model, we denote $\hat{p}(f(x))$ as the maximum score of an image among seen categories.
Thus, for an image from the seen domain, the $\hat{p}(f(x))$ should be extremely large in terms of the true label. Conversely, the $\hat{p}(f(x))$ should be small, indicating a uniform label distribution for an image from the unseen domain.
To this end, we compare the $\hat{p}(f(x))$ for all unseen domain samples by individually applying fully-supervised learning and our $\mathcal{L}_{ddc}$ to baseline DSP with noisy pseudo data.
The results are reported in Figure~\ref{fig:c}.

From Figure~\ref{fig:c}, it can be observed that, compared to fully supervised learning, $\mathcal{L}_{ddc}$ improves the percentage of samples with a small $\hat{p}(f(x))<0.5$ from $50\%$ to $70\%$ on CUB, $24\%$ to $42\%$ on AWA2, and $30\%$ to $50\%$ on aPY, approximatively.
On SUN, the percentage of samples with $\hat{p}(f(x))<0.3$ is improved from $22\%$ to $43\%$.
Notably, more samples with a small $\hat{p}(f(x))$ in Figure~\ref{fig:c} mean that more unseen domain samples can be distinguished from the seen domain samples.
Therefore, compared to fully supervised learning, $\mathcal{L}_{ddc}$ makes the embedding features between two domains more distinguishable, which accounts for our impressive performance.
Furthermore, it also shows that the softmax classifier $p$ can model the decision boundary between two domain.
Thus, using domain-specific classifiers are reasonable.

\noindent\textbf{Effects of varying $\tau$ values.}
$\tau$ is a critical parameter to judge whether a testing sample is from an unseen domain. The results of varying $\tau$ values are shown in Figure~\ref{fig:t}.
It can be found that, for the seen domain samples, a higher $\tau$ leads to a lower $MCA_s$.
The reason is that a higher $\tau$ may mistakenly give some seen domain samples to the ranking-based classifier, which degrades the $MCA_s$.
Conversely, in the unseen domain, the larger value the $\tau$ is, the higher $MCA_t$ our DSEN obtains. 
The reason is that a higher $\tau$ will feed a large number of unseen domain samples to the ranking-based classifier, which is good at unseen domain categorization and consequently improves the $MAC_t$.
As the metric $H$ is the combination of \textit{MCA}$_s$ and \textit{MCA}$_t$, $H$ does not have a consistent trend. 
With an increase of $\tau$, $H$ first increases to the optimal value and then drops. 
From Figure~\ref{fig:t}, it can be found that the optimal values for $\tau$ in different datasets are different, \emph{e.g.,} the optimal values for $\tau$ are 0.8, 0.5, 0.9, 0.8 for CUB, SUN, AWA2,and aPY, respectively.

\noindent\textbf{Effects of domain division constraint $\mathcal{L}_{ddc}$.}
We then verify the effectiveness of using domain division constraint $\mathcal{L}_{ddc}$. As shown in Table~\ref{tab:gzsl}, the baseline DDC obtains a higher harmonic mean ($H$) than S2V on all four datasets.
For example, on the AWA2 dataset, the DDC raises a harmonic mean ($H$) from $39.8\%$ to $60.1\%$ over S2V, which is mainly attributed to the significant $24.5\%$ improvement of $MAC_t$. Further, with the domain division constraint $\mathcal{L}_{ddc}$, the DSEN obtains a higher performance than DSP. 
These improvements confirm that the $\mathcal{L}_{ddc}$ can effectively make the embedding features more distinguishable, thereby rendering improved recognition in challenging unseen domains.

Furthermore, two domain-specific classifiers, based on $\mathcal{L}_{ddc}$, also yield significant contributions to our impressive performance.
As shown in Table~\ref{tab:gzsl}, the main improvements of our DDC come from the high $MAC_t$ in the unseen domain, which shows that the searching space reduction of the ranking-based classifier is an important factor of performance improvements.
Especially, on the CUB dataset, the metric H of DDC is 12.9\% higher than FGN \cite{Verma2018} in terms of H, where the FGN uses a single softmax classifier in two domains.
This also proves that using two domain-specific classifiers based on $\mathcal{L}_{ddc}$ is superior to using single shared classifier.

\begin{figure}
	\begin{center}
		\includegraphics[width=0.9\linewidth]{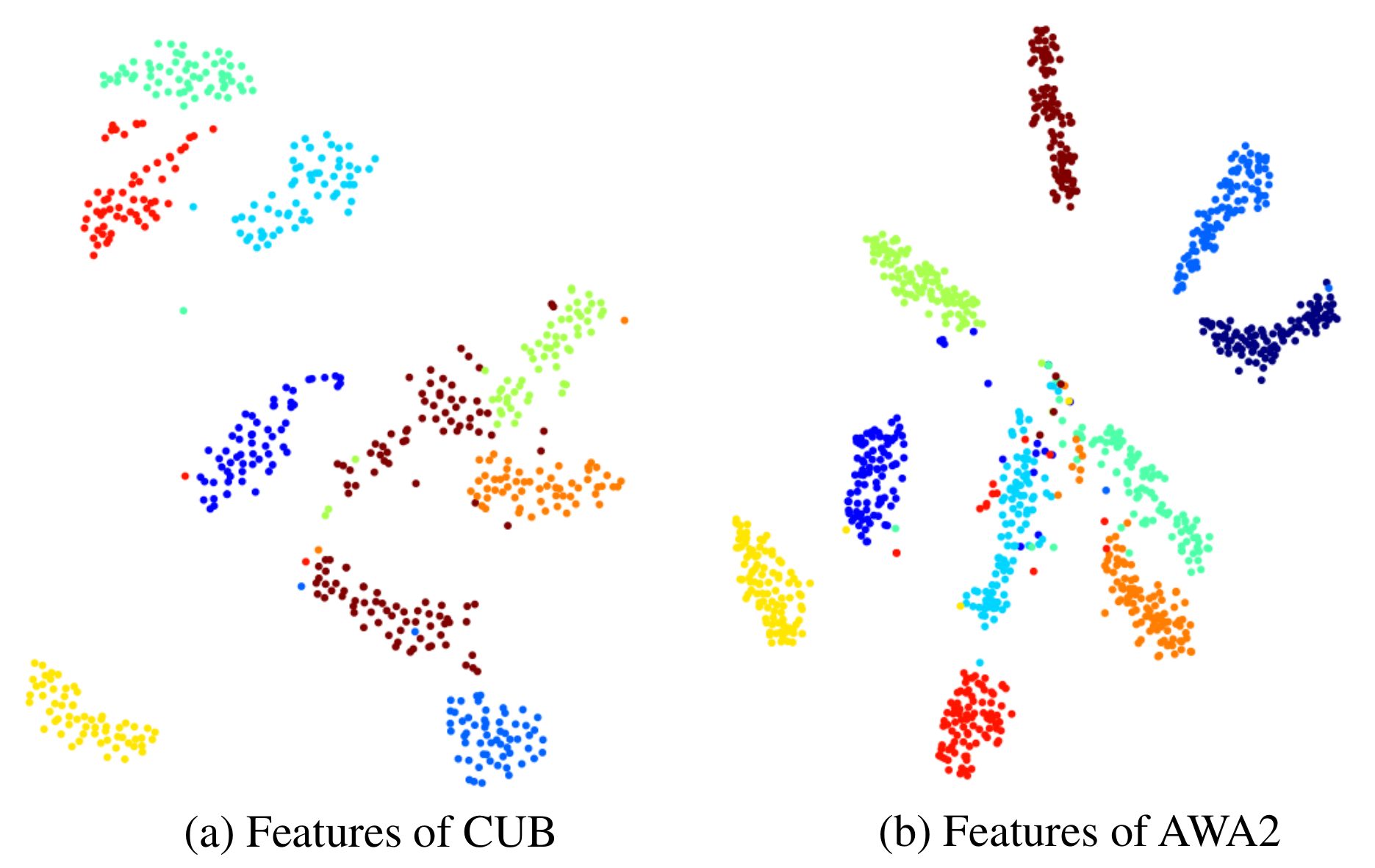}
	\end{center}
	\caption{The t-SNE of visual features from DSEN on CUB and AWA2, respectively. }
	\label{fig:tsne}
\end{figure}

\noindent\textbf{Feature visualizations of DSEN.}
Figure~\ref{fig:tsne} shows the t-SNE of generated visual features by DSEN on CUB and AWA2 datasets, respectively.
In each dataset, total $10$ categories are randomly selected from the unseen domain.
From the results, DSEN can not only preserve the semantic relationship in the embedding space but also obtain a large inter-class discrimination.
This is attributed to DSP that captures accurate domain difference and DDC that enlarges the domain difference.

\subsection{Comparison with existing methods}
\noindent\textbf{Comparison with generalized zero-shot learning.}
Table~\ref{tab:gzsl} illustrates comparison with previous methods on generalized ZSL. As shown in Table~\ref{tab:gzsl}, our DSEN significantly outperforms existing methods on four datasets, \emph{e.g.,} DSEN obtains $15.0\%$, $1\%$, $3.5\%$, and $16.8\%$ improvement in terms of metric $H$ on CUB, SUN, AWA2, and aPY, respectively.

\begin{table*}
\begin{center}
\caption{Evaluation performance under generalized zeros-shot learning. NG indicates non-generative methods, and G indicates generative methods that use GAN.} \label{tab:gzsl}
\begin{tabular}{c|l|c|c|c|c|c|c|c|c|c|c|c|c}
  \hline
&\multirow{2}{*}{Methods}&\multicolumn{3}{c|}{CUB~\cite{Welinder2010}}&\multicolumn{3}{c|}{SUN~\cite{Patterson2012}}&\multicolumn{3}{c|}{AWA2~\cite{Xian2018}}&\multicolumn{3}{c}{aPY~\cite{Farhadi2009}}\\
\cline{3-14}
&&\textit{MCA}$_t$&\textit{MCA}$_s$&$H$&\textit{MCA}$_t$&\textit{MCA}$_s$&$H$&\textit{MCA}$_t$&\textit{MCA}$_s$&$H$&\textit{MCA}$_t$&\textit{MCA}$_s$&$H$\\
\hline
\hline
\multirow{8}{*}{NG}&CMT\cite{Socher2013}&7.2&49.8&12.6&8.1&21.8&11.8&0.5&90.0&1.0&1.4&$\textbf{85.2}$&2.8\\
&SYNC\cite{Changpinyo2016}&11.5&70.9&19.8&7.9&$\textbf{43.3}$&13.4&10.0&90.5&18.0&7.4&66.3&13.3\\
&SAE\cite{kodirov2017semantic}&7.8&54.0&13.6&8.8&18.0&11.8&1.1&82.2&2.2&0.4&80.9&0.9\\
&KL\cite{Zhang2018}&19.9&52.5&28.9&19.8&29.1&23.6&17.6&80.9&29.0&11.9&76.3&20.5\\
&PTZSL\cite{long2018pseudo}&23.0&51.6&31.8&19.0&32.7&24.0&-&-&-&15.4&71.3&25.4\\
&CDL\cite{Jiang_2018_ECCV}&23.5&55.2&32.9&21.5&34.7&26.5&-&-&-&19.8&48.6&28.1\\
&PSR-ZSL\cite{Annadani2018}&24.6&54.3&33.9&20.8&37.2&26.7&20.7&73.8&32.2&13.5&51.4&21.4\\
&SP-AEN\cite{Chen2018}&34.7&70.6&46.6&24.9&38.6&30.3&23.3&$\textbf{90.9}$&37.1&13.7&63.4&22.6\\
\hline
\multirow{2}{*}{G}&SE-ZSL\cite{Verma2018}&41.5&53.3&46.7&40.9&30.5&34.9&$\textbf{58.3}$&68.1&62.8&-&-&-\\
&FGN\cite{xian2018feature}&43.7&57.7&49.7&$\textbf{42.6}$&36.6&39.4&-&-&-&-&-&-\\
\hline
\multirow{4}{*}{}&S2V&25.6&56.6&35.3&20.1&35.3&26.2&25.6&88.9&39.8&15.5&73.6&25.7\\
&DSP&30.8&62.7&41.3&30.0&40.3&34.4&31.2&87.9&46.1&18.1&73.1&29.0\\
&DDC&57.1&69.2&62.6&40.1&39.2&39.6&51.3&75.2&61.0&30.9&44.9&36.6\\
&DSEN&$\textbf{59.1}$&$\textbf{71.1}$&$\textbf{64.5}$&39.4&41.4&$\textbf{40.4}$&56.4&80.4&$\textbf{66.3}$&$\textbf{31.6}$&52.1&$\textbf{39.4}$\\
\hline

\end{tabular}
\end{center}
\end{table*}

\begin{table}
\begin{center}
\caption{Conventional zeros-shot learning. The \textit{MCA} ($\%$) metric is used for comparison.} \label{tab:czsl}
\begin{tabular}{lcccc}
  \hline
  Methods&CUB&SUN&AWA2&aPY \\
  \hline
  \hline
  CAV\cite{zhang2017learning}&52.1&61.7&65.8&-\\
  FGN\cite{xian2018feature}&61.5&62.1&-&-\\
  SE-ZSL\cite{Verma2018}&59.6&63.4&69.2&-\\
  PSR-ZSL\cite{Annadani2018}&56.0&61.4&63.8&38.4\\
  CDL\cite{Jiang_2018_ECCV}&54.5&63.6&-&43.0\\
  SP-AEN\cite{Chen2018}&55.4&59.2&58.5&24.1\\
  LDF\cite{li2018discriminative}&70.4&-&-&-\\
  \hline
  S2V&52.4&58.2&65.8&40.5\\
  DSP&56.2&62.6&69.1&41.7\\
  DDC&$\textbf{71.8}$&$\textbf{64.0}$&71.2&43.1\\
  DSEN&$\textbf{71.8}$&62.2&$\textbf{72.3}$&$\textbf{43.5}$\\
  \hline
\end{tabular}
\end{center}
\end{table}

To evaluate the effectiveness of domain-specific projections, we compare the DSP baseline with two representative methods~\cite{Annadani2018,Jiang_2018_ECCV} which both employ a single shared semantic-visual projection. From Table~\ref{tab:gzsl}, we see that the DSP baseline performs best on all four datasets in terms of metric $H$.
The high performance demonstrates the superiority of our domain-specific projections to the single shared semantic projection. Comparing with \cite{Annadani2018,Jiang_2018_ECCV}, the other advantage of DSEN is that it makes the visual features more discriminative. In this work, we define the domain shift degree as $|\textit{MCA}_s-\textit{MCA}_t|$.
As the CUB for examples, we find that the domain shift degree for PSR-ZSL \cite{Annadani2018} and CDL \cite{Jiang_2018_ECCV} are both larger than $30\%$. However, our DSEN only has a $12\%$ domain shift degree.
This low domain shift degree proves that our domain-specific projections can generate domain-robust embedding features.

Different from embedding-based PSR-ZSL and CDL, SE-ZSL \cite{xian2018feature} and FGN \cite{Verma2018} obtain state-of-the-art performance by alleviating the domain shift problem with synthetic visual data in an unseen domain. However, they all employ the widely used fully supervised learning that can degrade the recognition performance on real seen domain data, \emph{i.e.,} FGN \cite{Verma2018}  obtains a 10\%  drop of $\textit{MCA}_s$ with synthetic data.
Instead, the $\mathcal{L}_{ddc}$ used in our DDC can reduce the influence of noisy synthesized data.
For example, in CUB dataset, the DDC obtains \textit{MCA}$_{s}$ of 69.2\%, which is higher than the values of 55.7\% and 46.7\% for FGN and SE-ZSL, respectively.
As a consequence, the high \textit{MCA}$_t$ and \textit{MCA}$_s$ make DSEN obtain the highest $H$ among all datasets, which also demonstrates its effectiveness in generalized ZSL.

\noindent\textbf{Comparison with conventional zero-shot learning.}
Comparison with conventional ZSL setting is shown in Table~\ref{tab:czsl}, where the testing images only come from an unseen domain.
Notably, conventional ZSL setting is easier than generalized ZSL due to it ignores the searching space from the seen domain.
From Table~\ref{tab:czsl}, we can observe that the proposed DSEN obtains the best performance on four datasets. 
Also, the DDC has achieved higher performance than the existing methods on four datasets. 
It proves that the powerful and discriminative visual representations by the end-to-end trainable visual network are significant.
Furthermore, compared to the $MCA_t$ in Table~\ref{tab:gzsl} in generalized zero-shot learning, we have found that the four baselines all obtain higher performance. 
The reason is that the conventional ZSL know prior information for the testing images belonging to which domains, which mitigates the projection domain shift problem. 


\noindent\textbf{Discussion.}
As shown in Table~\ref{tab:gzsl}, DSEN achieves impressive improvement on CUB, aPY, and AWA2. However, it cannot obtain a consistent improvement on SUN dataset.
The reason is that too many categories in SUN make it hard to generate good visual features from semantic attributes of low dimensions.
More specifically, FGN uses GAN to generate synthetic visual features for an unseen domain, which is much more powerful than our two-layer generator.
Thus, the distance between two generators is hard to remedy with domain-specific projections and classifiers, since there is a total of 717 categories in SUN. 
However, our DSEN finally obtains a slightly higher $H$ value than FGN, due to an obviously higher $MAC_s$ in the seen domain, which ensures the robustness of DSEN.

\section{Conclusion}
With an aim to solve the domain shift problem in generalized zero-shot learning, we propose a novel Domain-Specific Embedding Network by applying specific projections to seen and unseen domains based on domain characteristics.
In contrast to existing methods using a single shared projection, we demonstrate that domain-specific projections can better capture domain similarities and differences, leading to more robust embedding features.
To avoid domain-separated embedding space, a semantic reconstruction constraint is designed by using semantic labels to associate two specific projections in a cycle consistency way.
Furthermore, a domain division constraint is developed to make the generated embedding features more distinguishable.
Experiments on four benchmarks demonstrate the effectiveness of the proposed method.

In the future, powerful generators will be explored to provide more reliable synthetic visual representations, \emph{e.g.,} GAN.
Also, domain-specific projection architectures will be explored by using autoML, which may yield further improvements.

\section{Acknowledgement}
This work is supported by the National Key Research and Development Program of China (2017YFC0820600), National Defense Science and Technology Fund for Distinguished Young Scholars (2017-JCJQ-ZQ-022), the National Nature Science Foundation of China (61525206,61771468,61622211,61620106009),the Youth Innovation Promotion Association Chinese Academy of Sciences (2017209), National Postdoctoral Programme for Innovative Talents (BX20180358), and the Fundamental Research Funds for the Central Universities (WK2100100030).

  \bibliographystyle{ACM-Reference-Format}
  \bibliography{zsl}
  
\end{document}